%% file: main.tex
\documentclass[10pt,letterpaper]{article}

\usepackage{cogsci}
\usepackage{amssymb}
\usepackage{graphicx}
\usepackage{algpseudocode}
\usepackage{algorithm}
\usepackage{amsmath}
\usepackage[normalem]{ulem}
\useunder{\uline}{\ul}{}
\usepackage{enumitem}
\setlist[itemize]{itemsep=0.5pt}

\cogscifinalcopy 

\usepackage{pslatex}
\usepackage{apacite}
\usepackage{float} 

\title{Understanding Multimodal Deep Neural Networks: A Concept Selection View}
 
\author{
    {\large \bf Chenming Shang (scm22@mails.tsinghua.edu.cn)} \\
    \AND
    {\large \bf Hengyuan Zhang (zhy22@mails.tsinghua.edu.cn)} \\
    \AND
    {\large \bf Hao Wen (wh22@mails.tsinghua.edu.cn)} \\
    \AND
    {\large \bf Yujiu Yang (yang.yujiu@sz.tsinghua.edu.cn)} \\
    \AND
    Tsinghua Shenzhen International Graduate School, Tsinghua University, Shenzhen, China.}

\begin{document}

\maketitle

\input{0_abstract}
\input{1_introduction}
\input{2_setup}
\input{3_method}
\input{4_experiments}
\input{5_conclusion}
\input{6_acknowledgements}

\nocite{losch2019interpretability}
\nocite{wang2023learning}
\nocite{wang2020chain}
\nocite{kim2018interpretability}
\nocite{ramaswamy2022overlooked}
\nocite{rauker2023toward}
\nocite{marconato2022glancenets}
\nocite{margeloiu2021concept}
\nocite{he2019fine}
\nocite{shwartz2017opening}
\nocite{losch2019interpretability}
\nocite{kazhdan2020now}
\nocite{lundberg2017unified}
\nocite{abid2022meaningfully}
\nocite{li2023multi}
\nocite{zhang2022fine}
\nocite{kong2022multitasking}

\bibliographystyle{apacite}

\setlength{\bibleftmargin}{.125in}
\setlength{\bibindent}{-\bibleftmargin}

\bibliography{CogSci_Template}

\end{document}

%% file: 0_abstract.tex
\begin{abstract}
The multimodal deep neural networks, represented by CLIP, have generated rich downstream applications owing to their excellent performance, thus making understanding the decision-making process of CLIP an essential research topic. Due to the complex structure and the massive pre-training data, it is often regarded as a black-box model that is too difficult to understand and interpret. Concept-based models map the black-box visual representations extracted by deep neural networks onto a set of human-understandable concepts and use the concepts to make predictions, enhancing the transparency of the decision-making process. However, these methods involve the datasets labeled with fine-grained attributes by expert knowledge, which incur high costs and introduce excessive human prior knowledge and bias. In this paper, we observe the long-tail distribution of concepts, based on which we propose a two-stage Concept Selection Model (CSM) to mine core concepts without introducing any human priors. The concept greedy rough selection algorithm is applied to extract head concepts, and then the concept mask fine selection method performs the extraction of core concepts. Experiments show that our approach achieves comparable performance to end-to-end black-box models, and human evaluation demonstrates that the concepts discovered by our method are interpretable and comprehensible for humans.

\end{abstract}

%% file: 1_introduction.tex
\section{Introduction}
\label{sec:introduction}

Deep neural networks (DNNs) \cite{lecun2015deep} have achieved unprecedented success in a wide range of machine learning tasks, including computer vision \cite{han2022survey}, natural language processing \cite{guo2024connecting}, and speech recognition \cite{radford2023robust}. However, due to their complex and deep structures, they are often regarded as black-box models \cite{castelvecchi2016can}, which are too difficult to understand and interpret. In fields that demand high levels of trustworthiness, such as medicine \cite{9428234}, education \cite{zhang2024questioncentric}, and finance \cite{ozbayoglu2020deep}, how humans understand the DNNs has become increasingly crucial. It indicates whether we can trust the DNNs' decisions, and how we can rectify the errors when the DNNs make mistakes. Making deep learning models more interpretable is a significant yet challenging research topic.

Recently, there has been rapid development in multimodal pre-trained models \cite{han2023survey}, among which Contrastive Language-Image Pre-Training (CLIP) \cite{radford2021learning} has achieved remarkable progress by employing contrastive learning to enable the shared representation space for vision and text. Specifically, CLIP consists of a text encoder and an image encoder, to extract textual and visual features, respectively. It leverages natural language as supervision for images and establishes the correspondence between images and text by maximizing the similarity between an image and its corresponding text description while minimizing the similarity between text descriptions from different images. Through such cross-modal learning and extensive training with large-scale data, CLIP can realize a range of interesting applications, for instance, serving as a backbone for image representation extraction which then be directly utilized for classification tasks \cite{kumar2022fine}, highlighting the importance of understanding the underlying mechanism and capability of CLIP.

\begin{figure*}[t]
    \centering
    \includegraphics[width=\linewidth]{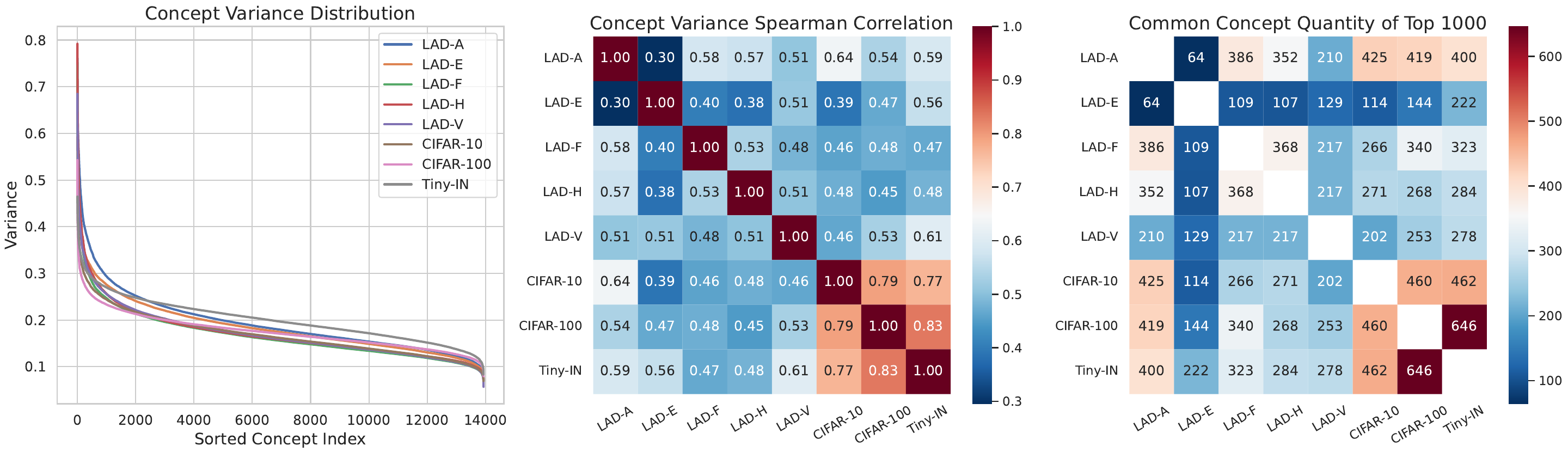}
    \caption{\textit{Left}: the distribution of the sorted concept variances. \textit{Middle}: the Spearman correlation coefficients of concept variances between any two datasets. \textit{Right}: the number of concepts shared in the top 1000 concepts with the highest variances between any two datasets.}
    \label{fig:observation}
\end{figure*}

DNNs can be interpreted at various levels, including pixels \cite{chattopadhay2018grad}, samples \cite{hammoudeh2022training}, weights \cite{wortsman2020supermasks}, individual neurons \cite{ghorbani2020neuron}, subnetworks \cite{amer2019review} and representations \cite{hendricks2018grounding}. However, with the increasing complexity of model structures, deeper layers, diverse data formats, and large-scale datasets, traditional model interpretability methods have become challenging to apply and generate human-understandable explanations. In this paper, we aim to address such a challenge: how to present the reasoning process of CLIP in a more intuitive manner and allow humans to intervene in the model's results?

A promising approach for achieving interpretability in deep learning is through concept-based models (CBMs) \cite{schwalbe2022concept,poeta2023conceptbased}, which map the visual representations to a set of human-generated high-level concepts to explain the black-box features of DNNs \cite{koh2020concept}. These interpretable concepts are then used to make the final decision by a linear function, greatly enhancing our understanding of the decision-making process. Due to the considerable cost of the fine-grained and precise annotation for each concept in CBMs, multimodal pre-trained model-based CBMs \cite{yuksekgonul2022post,oikarinen2023label} have recently emerged as a research hotspot. However, these recent innovations involve humans predefining a set of complex concepts specific to particular categories, such as \textit{small, black insect with six legs} \cite{yang2023language}, which introduces excessive human biases. It resembles generating a set of descriptions for images belonging to specific categories rather than truly understanding and explaining the decision-making process of CLIP.

To address the aforementioned challenges, we propose a two-stage Concept Selection Model (CSM) to understand concepts emerging from CLIP without introducing any human priors. Initially, we establish a concept library that CLIP can comprehend. We employ a powerful CLIP as a concept annotator to label the dataset with concepts, during which we observe the long-tail distribution of concepts. Subsequently, the concept greedy rough selection algorithm is applied to extract head concepts, and then the concept mask fine selection method performs the extraction of core concepts for specific classification tasks. Eventually, we conduct experiments using the filtered core concepts, which achieve comparable performance to end-to-end black-box models on multiple datasets. In addition, human evaluation demonstrates that the concepts discovered by our method are interpretable and comprehensible for humans.

%% file: 2_setup.tex
\section{Setup and Observation}
\label{sec:setup}

\subsection{Image Dataset and Concept Library}
We use the image classification task to understand the role of different concepts in the decision-making process of CLIP. For this purpose, we have chosen 2 common object datasets: CIFAR-10 and CIFAR-100 \cite{krizhevsky2009learning}, as well as a fine-grained object dataset: Large-scale Attribute Dataset (LAD) \cite{zhao2019large}, including 5 sub-datasets: LAD-Animal, LAD-Electronic, LAD-Food, LAD-Hair, and LAD-Vehicle. Additionally, the Tiny-ImageNet (Tiny-IN) dataset \cite{le2015tiny} is also employed as a reference to measure concept distribution.

\begin{figure}[t]
    \centering
    \includegraphics[width=\linewidth]{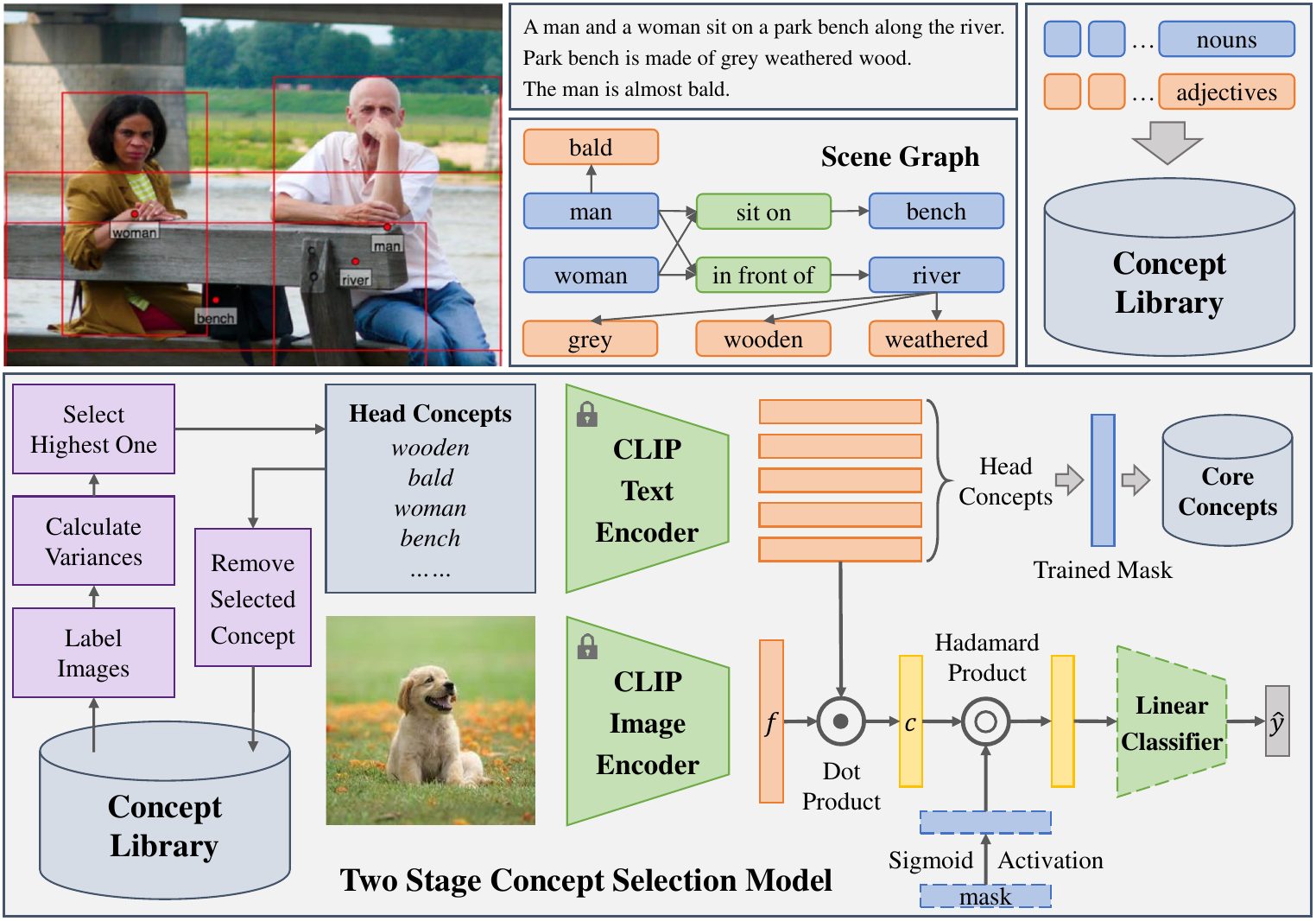}
    \caption{\textit{Up}: In the Visual Genome dataset, each image is accompanied by corresponding textual descriptions, which are transformed into scene graphs, with each word represented as an atomic node. \textit{Down}: Two-stage concept selection model: a rough selection is utilized to obtain the head concepts from the concept library, and subsequently a fine selection is applied to identify the core concepts from the head concepts.}
    \label{fig:framework}
\end{figure}

In order to enable the CLIP to automatically select and determine which concepts play the most important role in the decision-making process, thereby understanding the model's internal mechanisms instead of relying on pre-defined concepts by humans, which involve excessive human prior knowledge, we need to establish a concept library that satisfies the following properties: 
(\textit{i}).\textbf{Comprehensiveness}: the concept library is expected to encompass commonly used concepts in the open world, capturing a broad range of information and knowledge.
(\textit{ii}).\textbf{Atomicity}: each concept in the concept library should be atomic rather than composite, which is divided into the most basic semantic units.
We find that the scene graph consisting of visual concepts is an appropriate foundation. The scene graph derives from the Visual Genome dataset \cite{krishna2017visual} containing 100K images with corresponding text descriptions that provides a highly representative depiction of the real world, as shown in Fig.\ref{fig:framework}-Up. Meanwhile, in the scene graph, an atom is defined as an individual visual concept, corresponding to a single scene graph node, which ensures the principle of minimal semantic units.  Atoms are further subtyped into objects, relationships, and attributes. We pick the nouns and adjectives in them as the concept library with a quantity of 13,933.

\subsection{CLIP based Concept Annotator}
Consider a dataset of image-label pairs $\mathcal{D} = \{(x, y)\}$, where $x \in \mathcal{X}$ is the image and $y \in \mathcal{Y}$ is the label. We have $N$ concepts to describe the essential information of the world, which can be denoted as discrete tokens $\mathcal{E} = \{e_1,e_2,...,e_N\}$. Multimodal pre-trained alignment model (e.g., CLIP) has an image encoder $\Phi_I:\mathcal{X} \rightarrow \mathbb{R}^d$ and a text encoder $\Phi_T$, which can map images and text into a shared $d$-dimensional feature space respectively. We encode the discrete tokens with the CLIP text encoder then perform $L_2$ normalization\footnote{The symbol of $L_2$ normalization is omitted for convenience.} to obtain the concept embeddings $\{w_1,w_2,...,w_N | w_i = \Phi_T(e_i), i=1,2,...,N\}$ with the length of $1$ and the dimension of $d$. We concatenate these concept embeddings to a concept projection matrix $W_{N \times d}:\mathbb{R}^d \rightarrow \mathbb{R}^N$ in arbitrary order, also identified as a concept library.

Correspondingly, we utilize the CLIP image encoder to get the visual representations $f = \Phi_I(x)$. Considering that CLIP has aligned the images with the textual data when pre-training, the visual representations share the feature space with any concept embedding, and the projection length $\Vert f \Vert_2 \cdot \cos\langle w_i,f \rangle$ can reflect the presence of a particular concept in the image. Since undergoing $L_2$ normalization on $w_i$, the concept existence can be directly formulated in terms of the dot product $c=W \cdot f$. Thus far, we have utilized CLIP to annotate arbitrary concepts on any given image.

\subsection{Concept Distribution}
We utilize CLIP-ViT-L/14 to annotate 13,933 concepts on the training set images of 8 datasets. For each concept, we calculate its variance across different samples within the same dataset. A higher variance indicates a more notable difference in the presence of that concept across different images, suggesting that the concept may play a greater role in the task. We sort the concept variances and visualize the distribution of variances for each concept after ranking, as shown in Fig.\ref{fig:observation}-Left. We discover that different datasets exhibit remarkably similar distributions, all demonstrating a long-tail effect. Specifically, among the 13,933 concepts, only approximately 1,000 concepts have variances above 0.3 and 4,000 concepts have variances above 0.2, which implies that despite the numerous concepts available, the model relies on only a small fraction of them to make decisions.

We then analyze the fluctuations of these concepts across different datasets. In Fig.\ref{fig:observation}-Middle, we present the Spearman correlation coefficients of concept variances between any two datasets. We observe that all coefficients are positive, and the correlation coefficients between common object datasets are consistently above 0.77, indicating a strong positive correlation. This suggests that the model tends to depend on the same concepts when dealing with general image classification tasks. In addition, we find that the correlation coefficients between fine-grained object datasets are relatively smaller, which indicates that the model focuses on different concepts. The correlation coefficient reflects the similarity between datasets to a certain extent, for example, the coefficient between the LAD-Animal and LAD-Electronic is only 0.3, showing a lower correlation, which is consistent with the significant differences in images between these two datasets.

Furthermore, we present the number of concepts shared in the top 1000 concepts with the highest variances between any two datasets, reflecting the fluctuations of the head concepts, as shown in Fig.\ref{fig:observation}-Right. The head concepts exhibit a similar trend to the overall concepts, with more concepts shared among common object datasets and fewer shared among fine-grained object datasets. As the differences between datasets increase, the number of shared concepts decreases.

%% file: 3_method.tex
\section{Method: Concept Selection Model}
\label{sec:method}

From the observation, we conclude that only a few concepts produce major changes when reasoning in CLIP. Therefore, in this section, we propose a two-stage Concept Selection Model (CSM) to filter out the head concepts and further identify the core concepts.

\subsection{Greedy Rough Selection}

\begin{algorithm}[b]
\caption{Greedy Rough Concept Selection}
\begin{algorithmic}[1]
\State \textbf{Input:} $K$ images visual representations $V \in \mathbb{R}^{K \times d}$, $N$ concepts concept embeddings $W \in \mathbb{R}^{N \times d}$ identified as the concept library, head concept size $M$.
\State \textbf{Output:} head concept embeddings $W_{\mathrm{head}} \in \mathbb{R}^{M \times d}$.
\State \textbf{Initialization:} $W_{\mathrm{head}} \gets \emptyset$
\For{$i \in [1, \dots, M]$}
    \State $C \gets W \cdot V^{T}$
    \State $t^* \gets \arg\max_{t=1}^N \text{Var}(C[t])$
    \State $W_{\mathrm{head}} \gets W_{\mathrm{head}} \cup \{W[t^*]\}$
    \For{$k \in [1, \dots, K]$}
        \State $V[k] \gets V[k] - \cos\langle V[k], W[t^*]\rangle \cdot W[t^*]$
    \EndFor
\EndFor
\end{algorithmic}
\end{algorithm}

Selecting the optimal core concepts from a vast concept library is a combinatorial optimization problem, which becomes highly demanding when coupled with the optimization problem of linear classifiers. Therefore, based on the aforementioned observation, we adopt a greedy strategy to select the head concepts from the concept library in advance, narrowing down the scope of the concept library to avoid getting trapped in the local optima in the subsequent fine selection. 

\begin{table*}[t]
\centering
\caption{Comparison of different concept selection methods. Bold indicates the best and underline indicates the 2nd-best result.}
\vskip 0.12in
\begin{tabular}{ccccccccc}
\hline
Method         & CIFAR-10        & CIFAR-100       & LAD-A           & LAD-E           & LAD-F           & LAD-H           & LAD-V           & \textbf{Mean}   \\ \hline
Linear Probing & 0.9805          & 0.8703          & 0.9823          & 0.9397          & 0.8823          & 0.5176          & 0.9302          & 0.8718        \\
Random         & 0.7212          & 0.8323          & 0.8872          & 0.8269          & 0.7074          & {\ul 0.2482}    & 0.8102          & 0.7191          \\
Human Prior    & N/A             & N/A             & 0.9659          & 0.9049          & 0.7749          & 0.2145          & 0.8934          & N/A             \\
ConceptNet     & {\ul 0.8704}    & {\ul 0.8418}    & {\ul 0.9744}    & {\ul 0.9190}    & {\ul 0.8130}    & 0.1269          & \textbf{0.9172} & {\ul 0.7804}    \\
CSM            & \textbf{0.9708} & \textbf{0.8510} & \textbf{0.9767} & \textbf{0.9270} & \textbf{0.8516} & \textbf{0.4080} & {\ul 0.9129}    & \textbf{0.8426} \\ \hline
\end{tabular}
\label{tab:selection}
\end{table*}

Specifically, for a given dataset, we calculate the variance of the activation value of each CLIP-annotated concept across all images, which reflects the differences in a concept's presence, and we append the concept with the highest variance to the head concepts. An important consideration is that concepts with similar lexical meanings are likely to generate high variances, for example, \textit{dog} and \textit{puppy} could lead to concept redundancy, which we want to avoid. Therefore, after identifying the concept with the highest variance in each iteration, we need to eliminate its activation from all images to prevent the repetition of synonymous concepts. The details are depicted in Algorithm 1, where we pick the top $M=1000$ head concepts. Note that here we only utilize the image information from the training set without their labels, so the rough selection is weakly correlated with the specific downstream classification task and strongly related to the data itself.

\subsection{Mask Fine Selection}

After obtaining the head concepts, we proceed with the fine selection of core concepts. Specifically, we establish a learnable mask $m$, the size of which is equal to the number of head concepts, indicating the importance weights. Each weight in this mask is activated through a sigmoid function $\sigma(\cdot)$, mapping it to a value between 0 and 1, representing the significance of the respective concept, and values closer to 1 signifying higher importance. We perform a Hadamard product ($\odot$) of this importance weight with the concept activations generated by the head concepts, resulting in a weighted concept bottleneck. We then connect a linear classifier $\Psi_m:\mathbb{R}^M \rightarrow \mathcal{Y}$ for classification after this bottleneck, as illustrated in Fig.\ref{fig:framework}-Down, where we solve the following optimization problem:
\begin{gather}
    \hat{y} = \Psi_m \big( \sigma(m) \odot c_{\mathrm{head}} \big) = \Psi_m \Big( \sigma(m) \odot \big(W_{\mathrm{head}} \cdot \Phi_I(x) \big) \Big) \\
    \min_{\Psi, m} \mathop{\mathbb{E}}\limits_{(x,y) \in \mathcal{D}} \Big[ \mathcal{L} \big( \hat{y} ,y \big) \Big] + \lambda \cdot \Omega(\Psi_m)
\end{gather}
where $\mathcal{L}(\hat{y},y)$ is the cross-entropy loss function, $\Omega(\Psi)$ is a complexity measure, and $\lambda$ is the regularization strength.

We obtain the trained mask and sort it by magnitude, then select the top $N^*$ concepts as the core concepts $W_{\mathrm{core}}$. At next stage, we only use these core concepts to retrain the concept-based model by solving the following optimization problem:
\begin{equation}
    \min_{\Psi} \mathop{\mathbb{E}}\limits_{(x,y) \in \mathcal{D}} \Big[ \mathcal{L} \Big( \Psi \big(W_{\mathrm{core}} \cdot \Phi_I(x) \big),y \Big) \Big] + \lambda \cdot \Omega(\Psi)
\end{equation}
where $\Psi:\mathbb{R}^{N^*} \rightarrow \mathcal{Y}$ is a linear classifier. By following the steps, we have systematically filtered and obtained the top concepts as head concepts and the core concepts from the concept library. As a result, we have developed a concept-based model capable of performing classification tasks utilizing the core concepts.

%% file: 4_experiments.tex
\section{Experimental Results}
\label{sec:experimental}

In the comprehensive study of our approach (CSM) from the perspectives of \textbf{Accuracy (Q1-A1)} and \textbf{Interpretability (Q2-A2)}, we aim to answer the following questions:

\begin{itemize}
    \item \textbf{Q1.1}: What is an appropriate quantity of core concepts?
    \item \textbf{Q1.2}: Does CSM offer advantages compared to other concept selection methods?
    \item \textbf{Q1.3}: For what tasks do concept based models outperform black-box models?
    \item \textbf{Q2.1}: Which concepts are selected as core concepts?
    \item \textbf{Q2.2}: How do people understand and intervene in the model's reasoning and decision-making process?
    \item \textbf{Q2.3}: Does the concept based model actually function as we understand it in practice?
\end{itemize}

\begin{figure}[b]
    \centering
    \includegraphics[width=\linewidth]{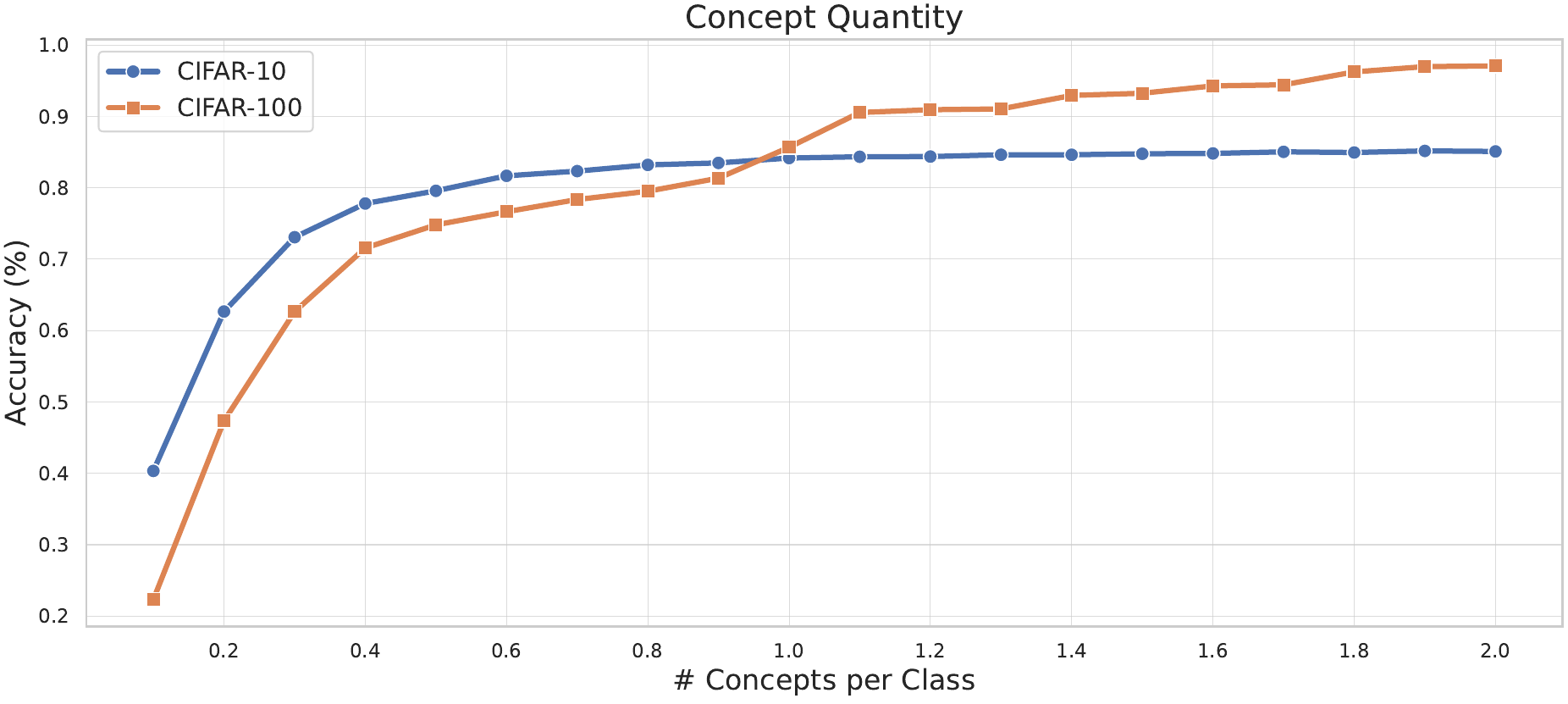}
    \caption{The variation in accuracy of the CSM as the concept quantity increases on CIFAR-10 and CIFAR-100.}
    \label{fig:quantity}
\end{figure}

\begin{figure*}[t]
    \centering
    \includegraphics[width=\linewidth]{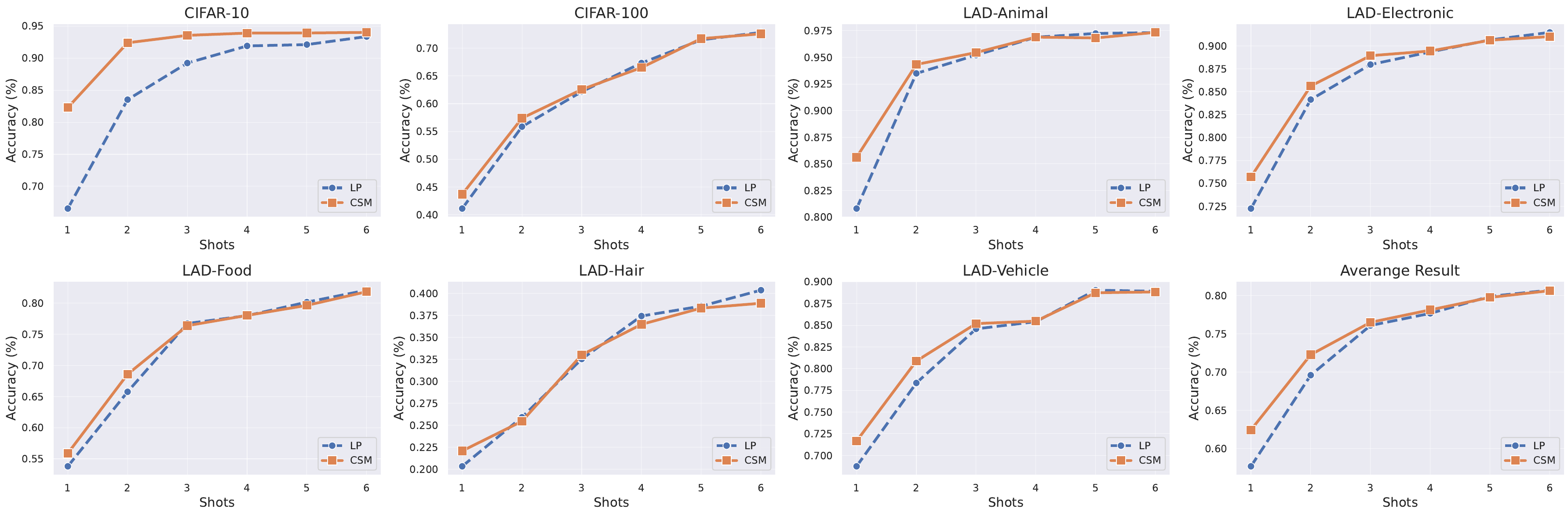}
    \caption{Accuracy comparison between CSM and linear probing. The x-axis indicates the number of labeled images per class.}
    \label{fig:fewshot}
\end{figure*}

\begin{table*}[t]
\centering
\caption{Top 5 core concepts with the highest activation variance for different datasets.}
\vskip 0.12in
\begin{tabular}{ccccccc}
\hline
CIFAR-10              & CIFAR-100             & LAD-Animal                   & LAD-Electronic             & LAD-Food             & LAD-Hair             & LAD-Vehicle                \\ \hline
\textit{fishing boat} & \textit{sundress}     & \textit{shepherd dog}   & \textit{range hood} & \textit{sandwich} & \textit{pile}     & \textit{streetcar}   \\
\textit{horseback}    & \textit{wildlife}     & \textit{pelican}        & \textit{lens cap} & \textit{bacon}    & \textit{braid}    & \textit{parasailing} \\
\textit{frog}         & \textit{streetcar}    & \textit{leopard}        & \textit{refrigerator}   & \textit{lettuce}  & \textit{ponytail} & \textit{cart}        \\
\textit{airplane}     & \textit{shepherd dog} & \textit{American bison} & \textit{monitor}    & \textit{tomato}   & \textit{hijab}    & \textit{minibike}    \\
\textit{yacht}        & \textit{tree}         & \textit{mermaid}        & \textit{earphone}  & \textit{walnut}   & \textit{barber}   & \textit{seaplane}    \\ \hline
\end{tabular}
\label{tab:top_concepts}
\end{table*}

\subsection{Accuracy}
\subsubsection{Concept Quantity (A1.1).} An intuitive judgment is that the number of concepts should be positively correlated with the number of categories. More categories require more concepts to support differentiation. Therefore, we conduct experiments on CIFAR-10 and CIFAR-100 according to the average number of concepts possessed by each category, as shown in Fig.\ref{fig:quantity}. Concept based model performs poorly when only a small number of concepts are available at the initial stage. As the number of concepts increases, the model's classification accuracy steadily improves, with a particularly noticeable improvement in the early stages when the concept quantity is small. When the number of concepts increased from approximately 0.1 concepts per class to 0.2 concepts per class, the accuracy on both CIFAR-10 and CIFAR-100 improved by more than 20\%. For CIFAR-10, the improvement becomes less pronounced at about 0.6 concepts per class for CIFAR-10 and at about 1.1 concepts per class for CIFAR-100. The model's accuracy approaches convergence at around 2 concepts per class. Therefore, in subsequent experiments, we choose the concept quantity as the category quantity times 2.

\subsubsection{Concept Selection Methods (A1.2).} We choose 3 other concept selection strategies as comparisons: (\textit{i}).\textbf{Random}: without distinguishing between good and bad concepts, an equal number of concepts to the CSM are selected directly and randomly from the concept library. (\textit{ii}).\textbf{Human prior}: for the LAD dataset, important concepts for the human task are manually identified. We similarly apply CLIP-ViT-L/14 for annotation to compare the performance of concept selection. (\textit{iii}).\textbf{ConceptNet}: ConceptNet is a knowledge graph dataset \cite{speer2017conceptnet}. We collect all concepts that have the \textit{hasA}, \textit{isA}, \textit{partOf}, \textit{HasProperty}, and \textit{MadeOf} relations with categories to select concepts. 

The results are shown in Tab.\ref{tab:selection}, where the uninterpretable black-box linear probing\footnote{Linear probing uses the representations extracted by CLIP image encoder to make predictions directly without interpretation.} is also presented for reference (not bolding). Our method achieves the highest classification accuracy on most datasets, even close to the unexplainable black-box linear probing, especially excelling on challenging tasks such as LAD-Hair. The random selection method serves as a baseline and performs the worst. Surprisingly, the human prior concept selection method does not perform well, implying a disparity between the model's reasoning and decision-making process and the human's process. It is worth mentioning that our method does not utilize any external knowledge, and all concept selection is done automatically by the model. Despite this, it outperforms the ConceptNet concept selection method, which incorporates human prior knowledge, highlighting the effectiveness of CSM concept selection.

\subsubsection{Few-shot Ability (A1.3).} The essence of the concept-based model is to decouple the representations of the uninterpretable black-box model into a set of meaningful concepts, in other words, to inform the model which information should be paid more attention to. When the amount of data is limited, concepts assist the model in extracting the crucial components relevant to the classification from the representation of the black-box model, and this distinct guidance helps the model focus more on the task, leading to superior performance. Fig.\ref{fig:fewshot} illustrates the performance comparison under different data volume settings between CSM and linear probing on 7 datasets and their average result. Compared to the black-box model, our method achieves superior performance when little data is available, and exhibits a slight performance gap with larger amounts of sample sizes, which indicates that our method has maintained accuracy without sacrificing interpretability. Meanwhile, our method does not introduce additional human priors, so the performance improvement cannot be attributed to external knowledge injection, but rather to better decoupling of knowledge inherent in the model.

\subsection{Interpretability}
\subsubsection{Core Concepts (A2.1).} Initially, we demonstrate the top 5 concepts with the highest activation value variances among the core concepts for each dataset, as shown in Tab.\ref{tab:top_concepts}. Consistent with our expectations, we observe that these concepts are tightly related to the core image compositions of the datasets, especially in fine-grained object datasets such as \textit{leopard} in LAD-Animal, \textit{monitor} in LAD-Electronic, \textit{walnut} in LAD-Food, \textit{ponytail} in LAD-Hair, and \textit{streetcar} in LAD-Vehicle. The fact that these concepts are understandable to humans suggests that CSM can capture concept variations that are consistent with human expectations.

\subsubsection{Model Debugging (A2.2).} 
\begin{figure}[t]
    \centering
    \includegraphics[width=\linewidth]{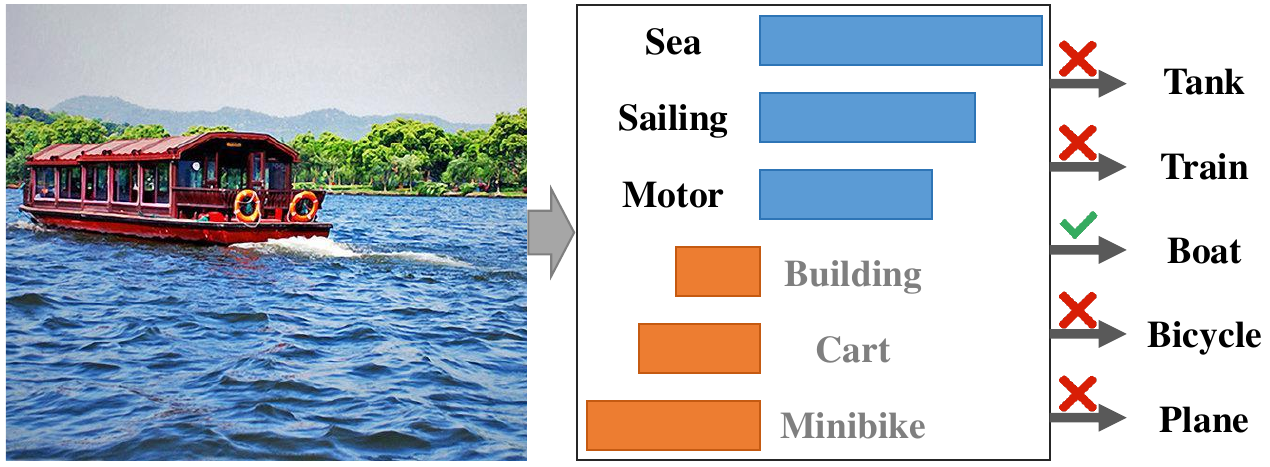}
    \caption{The decision process of the concept-based models.}
    \label{fig:interpretation}
\end{figure}

As an interpretable approach, one advantage of the concept-based model is its ability to help us understand the model's reasoning process, thereby making decisions more transparent. In Fig.\ref{fig:interpretation}, we present the top 3 concepts \textit{sea}, \textit{sailing}, and \textit{motor} with the highest normalized activation values in the CSM, as well as the bottom 3 activation concepts \textit{}{building}, \textit{cart}, and \textit{}{minibike}. The model's classification of the image as a \textit{boat} rather than a \textit{tank}, \textit{train}, \textit{bicycle}, or \textit{plane} is influenced by these concept activation values, which facilitates humans to comprehend the model's output.

\begin{figure}[t]
    \centering
    \includegraphics[width=\linewidth]{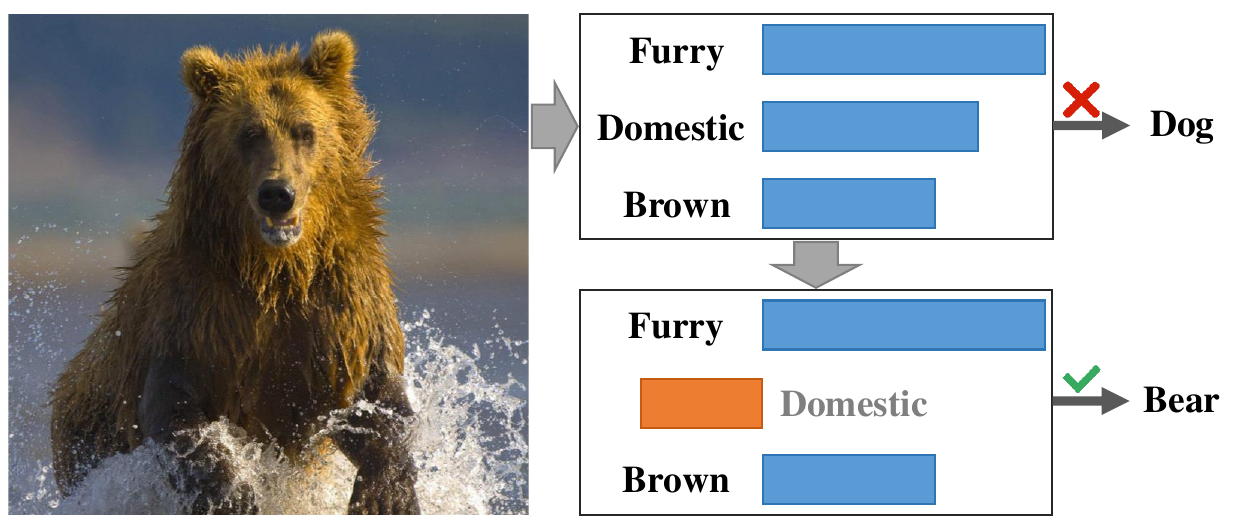}
    \caption{Model debugging through concept-based models.}
    \label{fig:debugging}
\end{figure}

Furthermore, after understanding the model's inference process, we can also intervene on the concepts of the misclassified samples to perform model debugging. For example, in Fig.6, the top 3 activation concepts are \textit{furry}, \textit{domestic}, and \textit{brown}, which lead the model to misclassify the image as a \textit{dog}. Upon analysis, it is determined that the concept \textit{domestic} is incorrect, resulting in the classification error. Therefore, by manually setting the activation value of that concept to 0 or a negative value, we can achieve model debugging and assist the model in correctly classifying it as a \textit{bear}.

\subsubsection{Human Evaluation (A2.3).} We evaluate the human interpretability in practical human-computer interactions through a user study and ablation of greedy rough selection (GRS) and mask fine selection (MFS) strategies. Specifically, we considered the following three scenarios:
(\textit{i}).\textbf{Image to Concepts (I2C)}: For a given image, we investigate whether humans perceive the concept with the highest CSM activation value. For each image, we display 1 concept from the top 5 highest activation values and randomly select 3 concepts from the remaining concepts, then ask users to determine which of the 4 concepts is most prominent in the image.
(\textit{ii}).\textbf{Concepts to Image (C2I)}: For a given combination of concepts, we assess whether humans can successfully predict the corresponding image. For each set of the top 5 highest concept activations, we display the corresponding image, another 1 image from the same category, and 2 additional images from different categories, then let users judge which of the 4 images best matches the concept activations.
(\textit{iii}).\textbf{Model Debugging (MD)}: When an image is misclassified, we examine whether humans could accurately identify the incorrect concept and correct it for model debugging. For a misclassified image, we provide the top 4 core concepts and prompt users to choose the concept most likely to be misidentified, then set the activation value of that concept to 0 and re-evaluate the accuracy.

\begin{table}[t]
\centering
\caption{User study results.}
\vskip 0.12in
\begin{tabular}{ccccc}
\hline
Acc. & random & w/o GRS     & w/o MFS     & CSM            \\ \hline
I2C  & 0.295  & 0.515       & {\ul 0.630} & \textbf{0.695} \\
C2I  & 0.320  & 0.570       & {\ul 0.645} & \textbf{0.680} \\
MD   & 0.080  & {\ul 0.260} & 0.215       & \textbf{0.330} \\ \hline
\end{tabular}
\label{tab:user_study}
\end{table}

For the user study, we randomly sample 5 categories from LAD-A and LAD-V, with 20 images per class. The study involves a total of 20 participants, and each answers 10 questions for every one of the above settings, the results are presented in Tab.\ref{tab:user_study}. Our method achieves the best results in all settings, as marked in bold. In tasks of concept-image correspondence, rough selection plays a crucial role, without which results in a severe drop in accuracy. In model debugging, fine selection is more pivotal, possibly due to the involvement of category information during mask training.

%% file: 5_conclusion.tex
\section{Conclusion}
\label{sec:conclusion}

By utilizing CLIP as a concept annotator, we observe the long-tail distribution of concepts, based on which we propose a concept selection model that explores concepts without introducing any human priors, enabling a deeper understanding of the reasoning process in multimodal DNNs. This concept-based decoupling method brings enhanced few-shot capability and permits applications such as model debugging. In the future, it will become a promising research topic to explore the nature of hierarchy and correlations between concepts, leveraging simple concept combinations to form higher-level concepts, and contributing to more transparent AI systems.

%% file: 6_acknowledgements.tex
\section*{Acknowledgements}

This work was partly supported by the National Natural Science Foundation of China (Grant No. 61991451) and the Shenzhen Science and Technology Program (JCYJ20220818101001004).